\documentclass{article}

 \usepackage[dblblindworkshop, preprint]{neurips_2025}
\usepackage{adjustbox}
\workshoptitle{Metacognition in Generative AI}

\usepackage[utf8]{inputenc} 
\usepackage[T1]{fontenc}    
\usepackage{hyperref}       
\usepackage{url}            
\usepackage{booktabs}       
\usepackage{amsfonts}       
\usepackage{nicefrac}       
\usepackage{microtype}      
\usepackage{xcolor}         

\usepackage{multirow} 
\usepackage{fancyvrb}
\usepackage{fvextra}
\usepackage{float}
\usepackage{amssymb}
\usepackage{makecell}
\usepackage{linguex}
\usepackage[dvipsnames]{xcolor}
\usepackage{graphicx}
\usepackage{array} 

\title{Evaluating Long-Term Memory for  Long-Context Question Answering}

\author{%
  Alessandra Terranova\thanks{Corresponding Author.},   
  Björn Ross, 
  Alexandra Birch \\
  School of Informatics, The University of Edinburgh\\
  \href{mailto:email@domain}{a.terranova@ed.ac.uk} \\
}

\begin{document}

\maketitle

\begin{abstract}
In order for large language models to achieve true conversational continuity and benefit from experiential learning, they need memory. While research has focused on the development of complex memory systems, it remains unclear which types of memory are most effective for long-context conversational tasks. We present a systematic evaluation of memory-augmented methods on long-context dialogues annotated for question-answering tasks that require diverse reasoning strategies.
We analyse full-context prompting, semantic memory through retrieval-augmented generation and agentic memory, episodic memory through in-context learning, and procedural memory through prompt optimization. Our findings show that memory-augmented approaches reduce token usage by over 90\% while maintaining competitive accuracy. Memory architecture complexity should scale with model capability, with foundation models benefitting most from RAG, and stronger instruction-tuned models gaining from episodic learning through reflections and more complex agentic semantic memory. In particular, episodic memory can help LLMs recognise the limits of their own knowledge.
\end{abstract}

\section{Introduction}

Memory is at the core of how humans think, learn, and make decisions. Similarly, for large language model (LLM) based agents, memory is fundamental to carry out extended conversations, reason over longer time frames, learn from experience, and act in a coherent way \cite{liu2025advances}.
LLMs often struggle to maintain coherence and accuracy in long, multi-turn interactions and very extended contexts. They also lack reliable mechanisms for metacognition, for instance they struggle with recognising the limits of their own knowledge or "knowing what they don't know" \cite{johnson2024imagining}. 

These challenges gain more importance as context windows become larger, conversations can span hundreds of turns,  and learning from experience becomes essential in real-world applications. 
Simply feeding models longer context is not always practical or efficient, as it increases inference cost and does not address the need for models to also learn from past information. 
Techniques such as retrieval-augmented generation (RAG) \cite{lewis2020retrieval} address context limitations and allow updating and controlling data sources to improve the factuality of generated text.
Additionally, recent advancements have introduced mechanisms to augment LLMs with memory components, to enhance their ability of retaining information over long contexts, and of learning from experience \cite{shinn2023reflexion, zhao2024expel}. 

In this paper, we examine different memory architectures, semantic, episodic (in-context learning with reflections), and procedural (prompt optimization), within a unified evaluation framework. We focus on long-context conversational QA, and reflect on how memory augmentation can support learning from experience and improve the models' ability to reason about their own knowledge.
Our findings show that memory-augmented approaches reduce token usage by over 90\% while maintaining competitive accuracy. The complexity of the memory architecture should scale with the model's capabilities: foundation models benefit most from RAG, while more advanced instruction-tuned models from episodic memory and richer semantic memory structures. Additionally, episodic memory plays an important role in mitigating LLMs' metacognitive limitations, helping them recognise the limits of their knowledge. This comparison establishes baselines and identifies trade-offs between performance, interpretability, and resource constraints that can inform memory system design.

\section{Background}
\subsection{Memory for LLM agents}
The need for a dedicated memory component in LLM-based agents seems fundamental to capture the model's internal reasoning, its task-specific context, historical dialogue, and evolving objectives \cite{liu2025advances, zhang2024survey}.
Similarly to human memory, memory modules for LLMs can be categorised into long and short-term memory.
Short-term memory -- usually the immediate context window of the LLM -- is sufficient to deal with simple tasks, but long-term memory becomes essential in more complex and realistic interaction scenarios \cite{khosla2023survey}.
Following the organization typically used for human memory, long-term memory is organised into three different mechanisms: \textbf{semantic memory} stores general world knowledge and facts, making it possible for the model to provide informed responses; \textbf{procedural memory} encodes knowledge about how to perform the task at hand; \textbf{episodic memory} finally captures specific past interactions or experiences, allowing the model to gather information that can inform its future decision making \cite{hatalis2023memory}.
Early approaches to memory through appending conversation history to the input prompt have evolved \cite{liu2025advances}, leading to the use of vector embeddings for retrieving memories and selective incorporation of reasoning steps into subsequent inference calls \cite{liu2023think}.
A growing body of work has been investigating dedicated memory systems for LLMs, exploring memory storage and management \cite{zhong2024memorybank, smolagents}. Some systems maintain logs of each agent interaction \cite{zhong2024memorybank}, while others employ read-write mechanisms that allow models to update memory content as needed \cite{modarressi2023ret}. 
Recent research has been exploring more dynamic and agent-driven approaches to memory. A-mem \cite{xu2025mem} proposes a semantic memory system where each memory is structured with contextual tags and dynamically linked to related memories. This enables the system to form an evolving network of knowledge that supports updates over time and memory reorganization. 

\subsection{Retrieval Augmented Generation}

Retrieval Augmented Generation (RAG) \cite{lewis2020retrieval} can improve LLMs efficiency and traceability, providing a way to capture and update knowledge in a modular and more interpretable way \cite{Guu_Lee_Tung_Pasupat_Chang_2020}. Retrieval-augmented LMs can be more reliable, adaptable, and attributable thanks to the use of large-scale datastores for knowledge-intensive tasks \cite{asai2024reliable}.
The standard RAG process involves chunking a set of documents, embedding and indexing them in a vectorstore, and retrieving relevant chunks based on semantic similarity with the query; this retrieved context can then be used to augment the LLM's prompt for generation.
Recent research has made these systems agentic by introducing more autonomy in determining what to retrieve and when, possibly having different retrievers available and refining the search strategies based on intermediate results \cite{shao2023enhancingretrievalaugmentedlargelanguage}.

\subsection{Learning through memory}
A growing interest has developed in leveraging LLM-based agents' experience to improve them without relying on costly fine-tuning or parametric updates. Traditional reinforcement learning approaches, while effective in many domains, often require extensive interaction data and prolonged training times. In response to these limitations, several frameworks have been proposed that enable language agents to learn through memory and linguistic feedback.
One such approach is Reflexion \cite{shinn2023reflexion}, which allows agents to self-improve through natural language reflection stored in an episodic memory buffer, rather than weight updates. 
Building on Reflexion, the Experiential Learning (ExpeL) framework \cite{zhao2024expel} proposes a memory-based strategy in which agents autonomously learn from experience without gradient updates, highlighting their capacity for transfer learning and emergent reasoning abilities.

Our approach to episodic memory, motivated by this literature, also makes use of textual feedback signals and knowledge from past experiences, but it is built with conversational  tasks in mind, and to be modular and easy to evaluate in conjunction with different LLM memory components.
These developments point to a paradigm shift in how learning is implemented in LLM-based agents, moving towards dynamic, introspective, and context-aware memory systems. 

\section{Methodologies}

\subsection{Dataset and evaluation}
Our goal is to test different memory strategies on a task which models realistic long-context conversations, which is why we decided to use LoCoMo (Long-term Conversational Memory) \cite{maharana2024evaluating}. LoCoMo is a publicly available evaluation benchmark consisting of very long-context synthetic conversational data. 
While synthetic data has limitations regarding real-world noise and ambiguous user intent, it provides controlled evaluation conditions that allow us to isolate memory mechanism effects without confounding variables from dialogue structure or inconsistent annotation quality. 
While previous datasets contain dialogues with around 1,000 tokens, spanning over 4-5 sessions, LoCoMo consists of ten conversations, spread across up to 35 chat sessions, each extending over 300 turns and averaging 9,000 tokens. The dataset is granularly annotated for question-answering requiring five distinct reasoning types: single-hop, multi-hop, temporal, open-domain or world knowledge, and adversarial.
In particular, temporal reasoning questions require the model to use the date of conversations, while adversarial questions do not have an answer in the given data, presenting both a challenge and possible insights into the trustworthiness of the examined systems. 
Additionally, we validate our results on QMSum \cite{zhong2021qmsum}, a human-annotated benchmark for query-based multi-domain meeting summarization. While the meetings are shorter and the queries do not require adversarial, temporal, or open-domain reasoning, this data can show the nuance of real-world conversations in product-design, academic, and parliamentary meetings.

For evaluation, we employ the F1 score to assess answer accuracy. As QA annotations are directly taken from the conversations as much as possible \cite{maharana2024evaluating}, we instruct the models to answer with exact words from the conversation.  As a measure of relative performance, we report the average F1 ranking across categories for each approach examined, and we use the average token length per query to measure efficiency.
Following Maharana et al. \cite{maharana2024evaluating}, for adversarial questions we set the evaluation score to 1 if the generated answer contains 'no information available' and to 0 otherwise. 
Unlike previous research that used different prompts depending on the type of question \cite{maharana2024evaluating,xu2025mem},  we prompt the models in the same way regardless of the question type, making the task more realistic and allowing better isolation of memory mechanism effects. 

\subsection{Memory for Question Answering}
We compare several approaches to memory under the QA setting. 
Table \ref{tab:mem_types} shows which types of memory are implemented by each approach.

\begin{table}[h]
\caption{\textbf{Comparison of memory strategies.} Each approach is categorised by memory types: short-term working memory (via context window), long-term memory (semantic, procedural, episodic), and whether it includes agentic control.}
        \footnotesize
        \centering
        \setlength{\tabcolsep}{2pt} 
        \begin{tabular}{@{}l*{5}{c}@{}} 
        \toprule
        & \textbf{Short-t.}& \multicolumn{3}{c}{\textbf{Long-term}} & \\
        \cmidrule(lr){2-2} \cmidrule(lr){3-5}
        \textbf{Strategy} & \rotatebox{90}{\small\textbf{Working}} & 
        \rotatebox{90}{\small\textbf{Semantic}} & 
        \rotatebox{90}{\small\textbf{Procedural}} & 
        \rotatebox{90}{\small\textbf{Episodic}} & 
        \rotatebox{90}{\small\textbf{Agentic}} \\
        \midrule
        Full Context    & \checkmark \\
        RAG             & \checkmark & \checkmark \\
        A-Mem           & \checkmark & \checkmark & & & \checkmark \\
        RAG + PromptOpt & \checkmark & \checkmark & \checkmark \\
        RAG + EpMem     & \checkmark & \checkmark & & \checkmark \\
        RAG + PromptOpt + EpMem& \checkmark & \checkmark & \checkmark & \checkmark \\
        \bottomrule
        \end{tabular}
        \label{tab:mem_types}
\end{table}

\paragraph{Full Context Prompting}
As a strong upper bound, we evaluate a naive approach in which the entire conversation history is appended to the prompt before the query. This allows the model full access to all past turns, eliminating the need for memory management or retrieval. 

\paragraph{RAG: Semantic Memory as Retrieval-Augmented Generation} 
We implement RAG using a top-$k$ retrieval pipeline over the full conversation history. Our RAG component represents semantic memory, as it stores the essential knowledge that grounds an agent's responses. 
At inference time, for each question, the model is provided with the top-$k$ relevant utterances and their respective timestamps, ranked by cosine similarity to the query using bge-m3 embeddings. We initially test $k = {5, 10, 20}$ to examine the impact of context window size on performance (Appendix A), and run all of our LoCoMo experiments with $k = 10$. The retrieved snippets are appended to the model prompt before answer generation.

\paragraph{A-Mem: Agentic Semantic Memory} We replicate the A-Mem architecture as introduced by Xu et al. \cite{xu2025mem}, where the system maintains structured memory notes representing each utterance in the conversational data. As the memory agent is presented with a new utterance for the same conversation, it updates its memory with new entries and can make decisions about updating old ones and their connections. For QA, top-10 semantic memory entries are retrieved and inserted into the prompt. We follow the authors’ implementation and prompting strategy, excluding task-specific rewording for adversarial and temporal questions.
Both RAG and A-Mem implement semantic memory, but with different levels of complexity. A-Mem supplements snippet embeddings with relevant keywords, contextual tags, and arrays of related memories. While RAG retrieves based purely on semantic similarity, A-Mem includes query expansion and provides the complete memory note context. 

\begin{figure*}[ht!]
\footnotesize
\centering
    \includegraphics[width=\linewidth, trim={0 25cm 0 0},clip]{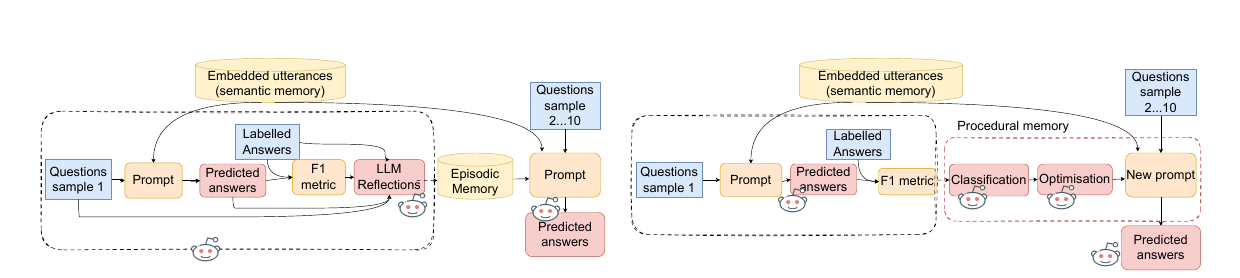}
    \caption{\textbf{Left figure: EpMem, episodic memory through in-context learning.}
    The model generates answers for a sample of questions and reflects on its performance using F1 scores and labelled answers to produce natural language reflections. These previous examples and reflections are stored in episodic memory and retrieved as in-context examples. \textbf{Right figure: PromptOpt, procedural memory through prompt optimization.}
    The model answers a sample of questions using an initial prompt and predictions are compared with labelled answers to compute F1 scores. A classification and optimization step is then used to update the prompt, forming procedural memory.}
    \label{fig:mem}
\end{figure*}

\paragraph{PromptOpt: Procedural Memory Through Prompt Optimization} Procedural memory encodes how an agent should behave and respond. We start with the same prompts used for the base RAG approach (Appendix B) that define the core agent behaviour and then evolve through feedback and experience. 
In particular, after generating answers for all the questions related to the first conversation, we record the model's responses, the correct labels, and its performance, and iteratively prompt the model with batches of 5 examples to use to refine its instructions, aiming to learn which approaches work best for different situations (Figure \ref{fig:mem}).
This optimization follows LangMem's implementation of procedural memory \cite{langmem2024} and it is achieved by a classification step, where the model is prompted to select which parts of its prompt caused errors, and an optimisation step where the model is prompted to generate an updated version of those prompt parts.
We use the optimized prompt for the next sets of questions.

\paragraph{EpMem: Episodic Memory Through In-Context Learning} Episodic memory preserves past interactions as learning examples that guide future behaviour. We implement it by first generating answers for all the questions related to the first conversation, noting the model's responses and the correct labels, and then prompting the model to reflect on each experience and generate a "reflection" string. We store dictionaries containing question, prediction, label, and reflection and we retrieve the top-3 most similar experiences to use as in-context examples whenever answering a new question  (Figure \ref{fig:mem}). The structure and content of the prompts used is found in Appendix B.

\subsection{Implementation details}
Appendix B details all the prompts we employed for our experiments.
We deploy LLama 3.2-3B, Qwen2.5-7B, and the respective instruction-tuned versions through huggingface, we access GPT-4o mini through the official OpenAI API. We report our main findings in Tables \ref{tab:memory_results} and\ref{tab:f1_rankings}, and the full experimental results in Appendix \ref{sec:results}.
We conduct all of our experiments on a 24 GB GPU machine for 5 runs, and report the mean and standard deviation.
For the retrieval components, we employ k=10 for snippet and A-Mem memories selection with LoCoMo data and k=20 with QMSum. For episodic memories we limit the number of examples to 3. We use bge-m3 text embeddings for our approaches and the all-minilm-l6-v2 model used by the authors for A-Mem. We report hyperparameter details in Appendix A.
The results on GPT-4o mini validate our findings on a widely-used commercial model. Our focus on open weights smaller models, instead, serves an important practical purpose, as these represent the scale most practitioners can deploy in resource-constrained environments. 

\section{Experiments and Results}
In our empirical evaluation, we compare the effectiveness of various memory augmentation strategies across multiple language models. The task spans five reasoning categories, with different challenges and considerations for each of them: Single-Hop, Multi-Hop, Temporal, Open-Domain and Adversarial.
We evaluate the Full Context baseline
against different combinations of memory mechanisms: RAG, A-Mem, RAG+PomptOpt, RAG+EpMem, and RAG+PromptOpt+EpMem. 
In Table \ref{tab:memory_results}, we report the mean and standard deviation for average F1 across categories and the average number of tokens used per query as a proxy for efficiency. We calculate the average F1 ranking across categories, ordering the approaches by their F1 scores, from highest to lowest within each model (rank 1 to the highest F1 score, rank 2 to the next, and so on).
As we use the first data sample (10\% of the dataset) to generate episodic and procedural memories, we report experimental results on the remaining 9 samples (1787 data points out of 1986). While in a real-world system these memories would be generated along all conversations, fixing a "training" sample and using the rest for evaluation allows us to capture a snapshot of model behaviour and learning.

\begin{table*}[h!]
    \centering
 \caption{\textbf{Performance comparison of memory augmentation approaches across instruction-tuned language models and reasoning categories on LoCoMo.} We report mean F1 scores $\pm$ SD, rankings and average number of tokens used per query for each approach across five reasoning categories. Best results per model in bold.} 

    \begin{adjustbox}{max width=\textwidth}
    \setlength{\tabcolsep}{3pt}
\renewcommand{\arraystretch}{0.9}
    \begin{tabular}{@{}>{\raggedright\arraybackslash}p{0.02\linewidth}lrrrrrrr@{}}
    \toprule
    & \textbf{Approach} & \multicolumn{5}{c}{\textbf{Category}} & \multicolumn{2}{c@{}}{\textbf{Average}} \\
    \cmidrule(lr){3-7} \cmidrule(l){8-9}
    & & \textbf{Single-Hop}& \textbf{Multi-Hop}& \textbf{Temporal}& \textbf{Open Domain}& \textbf{Adversarial}& \textbf{F1 R.}& \textbf{Tokens}\\
    \midrule
\multirow{7}{*}{\rotatebox[origin=c]{90}{\parbox{1.4cm}{\centering \textbf{Llama}}}}
& Full Context & \textbf{32.45} $\pm$ 1.81 & \textbf{27.14} $\pm$ 0.89 & \textbf{12.62} $\pm$ 1.19 & \textbf{39.44} $\pm$ 0.29 & 25.92 $\pm$ 1.28 & \textbf{1.00} & 23265.98 \\
& RAG & 6.45 $\pm$ 0.90 & 8.22 $\pm$ 1.07 & 5.03 $\pm$ 1.91 & 5.65 $\pm$ 0.30 & 14.30 $\pm$ 2.56 & 4.50 & 658.11 \\
& A-Mem & 2.12 $\pm$ 0.82 & 2.17 $\pm$ 1.21 & 3.60 $\pm$ 0.85 & 2.51 $\pm$ 0.86 & 65.28 $\pm$ 8.25 & 4.00 & 2480.37 \\
& RAG+PromptOpt & 10.25 $\pm$ 0.56 & 9.65 $\pm$ 0.56 & 5.92 $\pm$ 0.83 & 8.34 $\pm$ 0.53 & 32.03 $\pm$ 2.00 & 2.83 & 821.42 \\
& RAG+EpMem & 19.35 $\pm$ 0.76 & 10.62 $\pm$ 0.83 & 4.81 $\pm$ 0.52 & 11.52 $\pm$ 1.15 & \textbf{71.84} $\pm$ 3.06 & 1.83 & 1305.82 \\
& RAG+PromptOpt+EpMem & 18.89 $\pm$ 1.61 & 14.84 $\pm$ 3.00 & 4.79 $\pm$ 0.54 & 16.31 $\pm$ 3.20 & 70.12 $\pm$ 1.55 & 2.83 & 1377.53 \\
\midrule

\multirow{7}{*}{\rotatebox[origin=c]{90}{\parbox{1.4cm}{\centering \textbf{Qwen}}}}
& Full Context & 10.47 $\pm$ 0.53 & 3.18 $\pm$ 0.22 & 3.76 $\pm$ 0.35 & 8.10 $\pm$ 0.15 & 77.53 $\pm$ 1.11 & 5.40 & 23376.87 \\
& RAG & \textbf{21.12} $\pm$ 0.54 & 8.25 $\pm$ 0.23 & 4.93 $\pm$ 0.25 & 13.93 $\pm$ 0.58 & 95.29 $\pm$ 0.27 & \textbf{2.20} & 695.74 \\
& A-Mem & 12.13 $\pm$ 0.99 & \textbf{12.08} $\pm$ 2.88 & \textbf{6.23} $\pm$ 0.42 & \textbf{17.77} $\pm$ 3.37 & 81.07 $\pm$ 9.06 & \textbf{2.20} & 3307.27 \\
& RAG+PromptOpt & 7.87 $\pm$ 0.13 & 2.05 $\pm$ 0.16 & 4.41 $\pm$ 0.71 & 9.66 $\pm$ 0.39 & 71.98 $\pm$ 1.60 & 5.60 & 1024.88 \\
& RAG+EpMem & 18.02 $\pm$ 0.53 & 6.48 $\pm$ 0.08 & 5.48 $\pm$ 0.29 & 9.82 $\pm$ 0.28 & \textbf{95.59} $\pm$ 0.74 & 2.80 & 1452.88 \\
& RAG+PromptOpt+EpMem & 19.14 $\pm$ 1.78 & 7.06 $\pm$ 0.51 & 5.39 $\pm$ 0.27 & 10.24 $\pm$ 0.27 & 93.93 $\pm$ 0.72 & 2.80 & 1453.28 \\
\midrule

\multirow{7}{*}{\rotatebox[origin=c]{90}{\parbox{1.5cm}{\centering \textbf{GPT}}}}
& Full Context & 31.79 $\pm$ 0.14 & 19.87 $\pm$ 0.64 & 12.19 $\pm$ 1.00 & \textbf{55.90} $\pm$ 0.76 & 52.94 $\pm$ 1.73 & 3.40 & 23132.49 \\
& RAG & 29.98 $\pm$ 0.71 & 31.75 $\pm$ 1.47 & 11.59 $\pm$ 1.29 & 49.41 $\pm$ 0.44 & 84.28 $\pm$ 0.19 & 3.20 & 649.17 \\
& A-Mem & 24.54 $\pm$ 0.41 & 25.65 $\pm$ 0.46 & 8.06 $\pm$ 0.93 & 37.39 $\pm$ 0.44 & 64.93 $\pm$ 0.54 & 4.80 & 3514.45 \\
& RAG+PromptOpt & 15.55 $\pm$ 0.29 & 13.17 $\pm$ 0.29 & 6.53 $\pm$ 0.42 & 22.76 $\pm$ 0.15 & \textbf{93.08} $\pm$ 0.29 & 5.00 & 668.90 \\
& RAG+EpMem & \textbf{32.21} $\pm$ 0.46 & 40.73 $\pm$ 0.67 & \textbf{12.20} $\pm$ 0.39 & 51.94 $\pm$ 0.67 & 77.64 $\pm$ 0.60 & \textbf{1.80} & 969.26 \\
& RAG+PromptOpt+EpMem & 32.05 $\pm$ 0.81 & \textbf{40.97} $\pm$ 0.74 & 11.28 $\pm$ 3.09 & 51.75 $\pm$ 0.31 & 76.61 $\pm$ 2.20 & 2.80 & 972.93 \\
\bottomrule
    \end{tabular}
    \label{tab:memory_results}
    \end{adjustbox}
\end{table*}

\subsection{Overall Performance}
The results in Table \ref{tab:memory_results} show significant variance in performance across different memory augmentation strategies. Retrieval and memory augmented generation approaches improve performance across most models, with instruction-tuned variants benefitting the most (Table \ref{tab:f1_rankings}). In most cases memory and retrieval-based approaches achieve competitive or superior F1 scores to the Full Context baseline, while using significantly fewer tokens.

RAG outperforms other approaches at Multi-Hop reasoning, suggesting that direct retrieval of similar utterances provides useful context for multi-step reasoning.
Temporal-reasoning, adversarial and open-domain knowledge questions were the most challenging settings, with differences in performance based on the approach used. 
LLMs struggle to understand time concepts within dialogues, which is consistent with findings on single-turn-based benchmarks and temporal grounding \cite{qiu2023large, wang2023tram}.
 While this effect is somewhat less noticeable in GPT-4o mini, LLMs' performance for open-domain QA often degrades in memory-augmented settings. 
 
The adversarial questions, approached with the same prompt used for all categories rather than being rephrased as MCQs, show how the Full Context approach is prone to mistakes, while memory-augmented methods provide more trustworthy answers but can over-generate "no information available" answers and degrade performance in other categories for less powerful models (6.48 $\pm$ 0.08 F1 for multi-hop vs. 95.59 $\pm$ 0.74 F1 for adversarial for Qwen with RAG+Ep Mem). 
The way adversarial questions are designed in LoCoMo often references something that has high surface-level semantic similarity with part of the conversation the question relates to (Figure \ref{fig:adv_ex}). For this reason, even in retrieval-augmented and memory-augmented settings the models will still have some confounding information in their context, making it particularly challenging to give the correct "No information available" answer, unless the model has strong instruction-following capabilities (i.e. results from GPT-4o-mini). 

We validate our findings on real-world conversations, and our results on the test split of QMSum (Table \ref{tab:qmsum}) show that our methods transfer to other long-context conversational domains and to real-world, noisy dialogue data. 
As the QMSum meetings are not as long-context as LoCoMo conversations (on average the input tokens for the Full Context approach are 13k vs 23k tokens) and do not contain adversarial and temporal questions, stronger models that can handle a long context like GPT-4o mini achieve the best performance with the full context. 
Interestingly, we observe that PromptOpt gains importance and boosts performance when adapting the prompts used for QA to a query-based summarisation task.
Episodic memory also strengthens performance: while all meetings have different context, the queries about them follow the same annotation schema, so past experiences can help the model answer correctly.

\subsection{Baseline Performance}
The Full Context performance is consistent with findings from Maharana et al. \cite{maharana2024evaluating}, showing that most long context LLMs can comprehend longer narratives but are prone to mistakes in Adversarial questions and can struggle with Multi-Hop queries when presented with extremely long contexts. This could be due to the "lost in the middle" effect, where models oversample from the beginning and end of their context window \cite{liu2023lost}. 

\begin{table}[h]
    \centering
    \begin{minipage}[t]{0.48\textwidth}
        \centering
        \footnotesize
\caption{\textbf{Performance comparison of memory augmentation approaches across instruct models on QMSum test set.} We report mean F1 scores. Best results per model in bold.} 
\begin{tabular}{>{\raggedright\arraybackslash}p{0.31\linewidth}lll}
\toprule
                   \textbf{Approach} & \textbf{Llama} & \textbf{Qwen}  & \textbf{GPT}    \\ 
\midrule
Full Context        & 10.71          & 21.96          & \textbf{27.70} \\
RAG                 & 8.88           & 20.94          & 20.03          \\
RAG+PromptOpt       & \textbf{20.45} & 23.32          & 21.78          \\
RAG+EpMem           & 18.86          & \textbf{23.99} & 24.16          \\
RAG+PromptOpt +EpMem& 19.19          & 23.97          & 23.89          \\ 
\bottomrule
\end{tabular}
\label{tab:qmsum}
    \end{minipage}%
    \hfill
    \begin{minipage}[t]{0.48\textwidth}
        \centering
        \caption{\textbf{Average F1 rankings comparison.} Base vs instruction-tuned models (lower is better)}
        \footnotesize
        \setlength{\tabcolsep}{2.5pt}
        \renewcommand{\arraystretch}{0.9}
        \begin{tabular}{@{}>{\raggedright\arraybackslash}p{0.31\linewidth}rrrr@{}}
        \toprule
        \multirow{2}{*}{\textbf{Approach}} & \multicolumn{2}{c}{\textbf{Llama}}  & \multicolumn{2}{c}{\textbf{Qwen}} \\
        \cmidrule(lr){2-3} \cmidrule(lr){4-5}
        & \textbf{Base} & \textbf{Inst.} & \textbf{Base} & \textbf{Inst.} \\
        \midrule
        Full Context & 2.00& 1.00& 2.33& 4.50\\
        RAG & 2.16& 4.50& 1.16& 2.00\\
        A-Mem & 3.33& 4.00& 3.16& 2.00\\
        RAG+PromptOpt & 3.16& 2.83& 3.50& 4.66\\
        RAG+EpMem & 3.50& 1.83& 3.83& 2.00\\
        \makecell[l]{RAG+PromptOpt\\+EpMem}& 3.33& 2.83& 3.50& 2.66\\
        \bottomrule
        \end{tabular}
        \label{tab:f1_rankings}
    \end{minipage}
\end{table}

\subsection{RAG vs. A-Mem}
Both performing retrieval over embedded and indexed utterances and using A-Mem can serve as the semantic memory or data source for long-context and knowledge-intensive QA.
Importantly, RAG is much more efficient and scalable than A-Mem, not just in terms of tokens used at inference time, but also due to the nature of the information being stored. Each utterance is stored as a separate memory in A-Mem, and this requires two LLM calls per utterance to process the memory and generate all the required meta-data, in addition to embedding and indexing.
Due to the requirement for consistent structured output, A-Mem performs poorly on foundation models. RAG shows substantially better performance across all foundation models. 
This gap narrows for instruction-tuned models, which are able to take advantage of the full potential of A-Mem. While RAG still outperforms A-Mem in GPT-4o mini, Llama 3B Instruct with A-Mem achieves an average F1 ranking of 4.0, compared to 4.5 for RAG.

\subsection{Impact of Episodic Memory}
The integration of Episodic Memory through ICL with RAG shows promising results. Rag+EpMem outperforms RAG on GPT-4o mini and Llama3.2 3B Instruct (respectively, 1.30 vs. 3.20 and 1.83 vs. 4.5 F1 ranking ), while it achieves better results than all other approaches but slightly worse than RAG on Qwen2.5 7B  Instruct. This suggests that incorporating episodic memories containing examples of model performance and reflections can provide valuable guidance for answering similar questions.
Improvements can be seen in the adversarial and temporal reasoning categories for Llama and Qwen Instruct, and in multi-hop and temporal reasoning queries for GPT-4o mini. This suggests that the reflections on failures that required a specific type of reasoning were useful for the model to infer how to respond to new queries of the same category.

\begin{figure}[h]
    \centering
    \setlength{\abovecaptionskip}{0pt}
    \setlength{\belowcaptionskip}{0pt}
    \begin{minipage}[t]{0.45\textwidth}
        \centering
        \includegraphics[width=\linewidth, trim={0 14cm 0 0},clip]{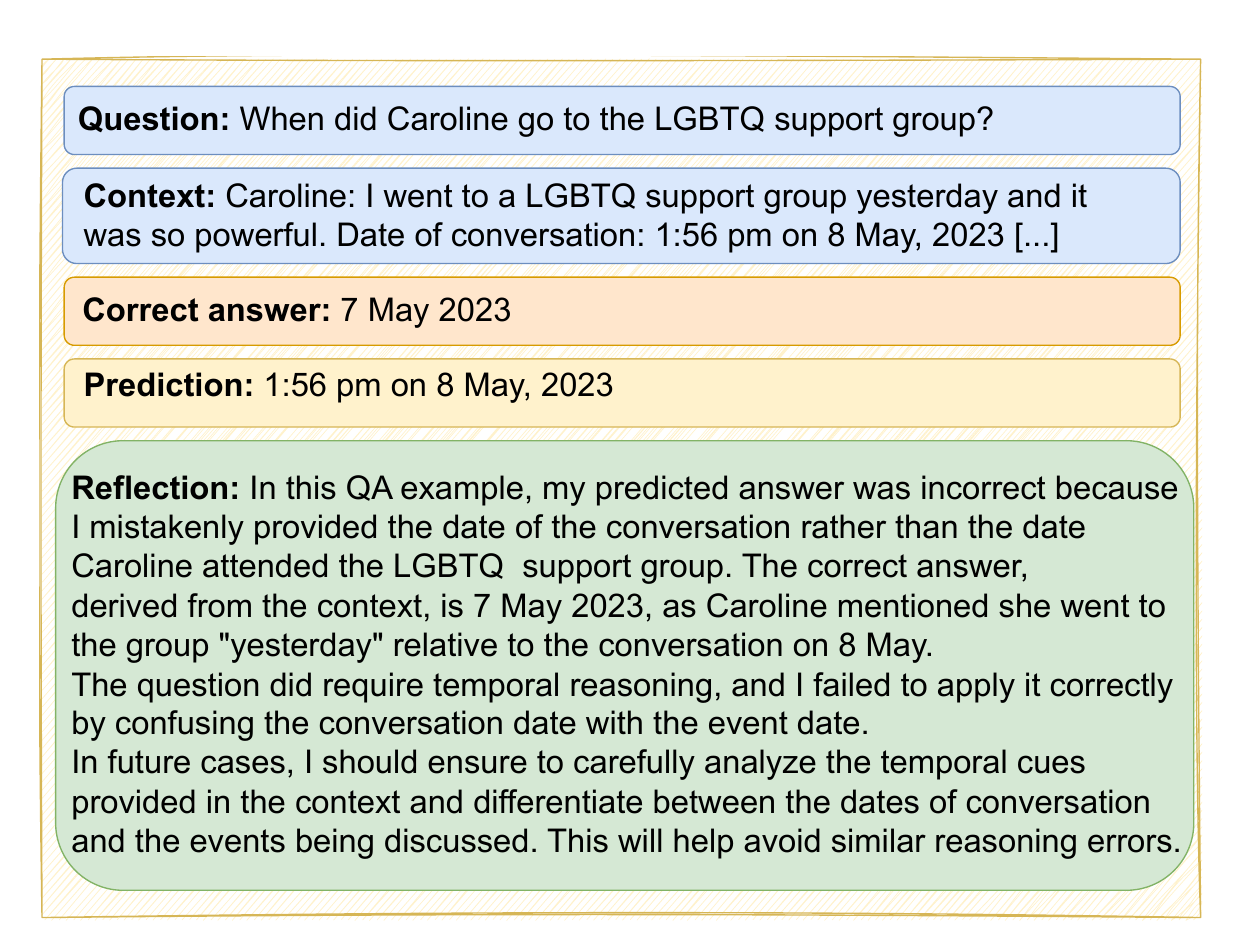}
        \caption{Example of episodic memory generated by GPT-4o mini}
        \label{fig:ep_ex}
    \end{minipage}
    \hfill
    \begin{minipage}[t]{0.45\textwidth}
        \centering
        \includegraphics[width=\linewidth, trim={0 14cm 0 0},clip]{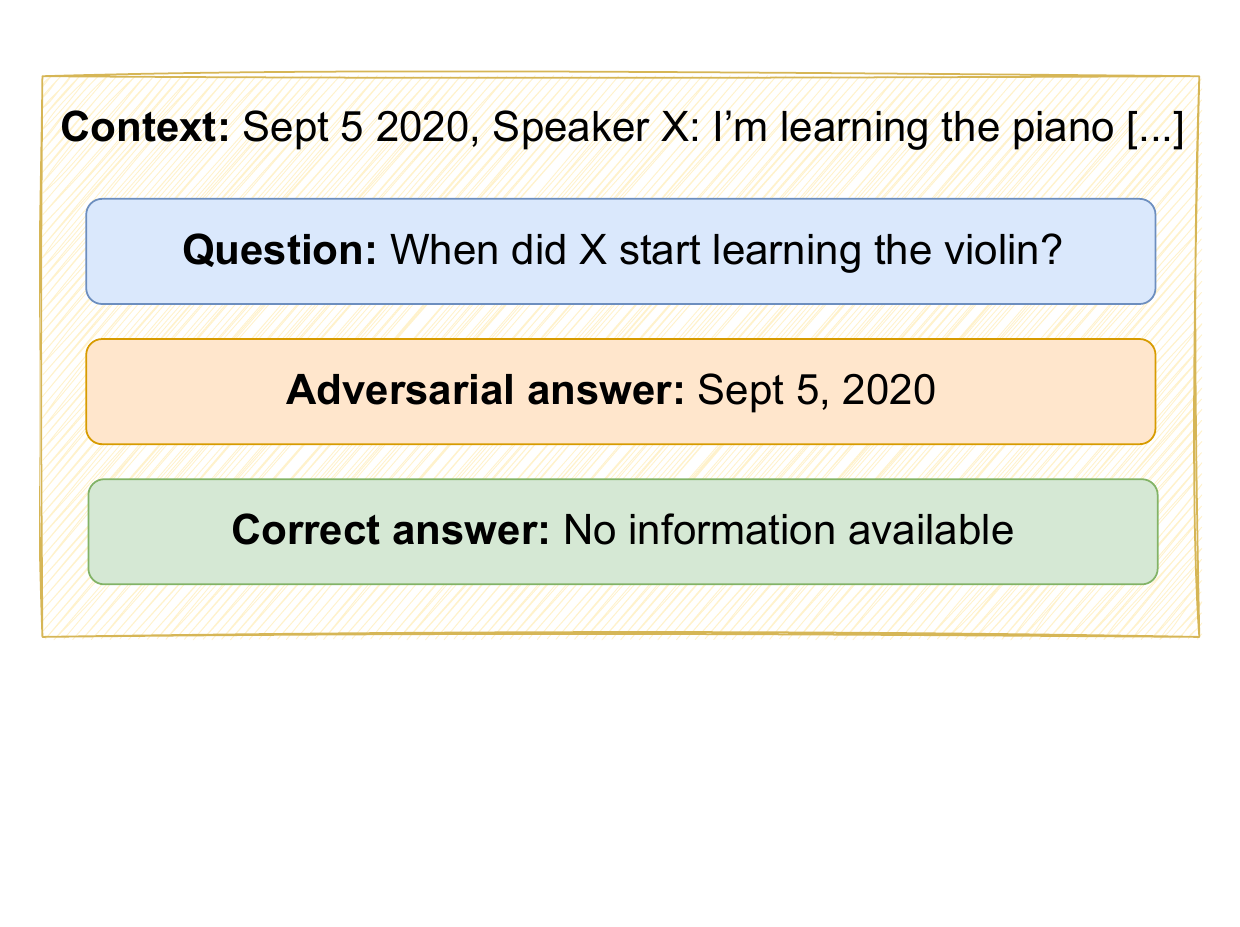}
        \caption{Example of adversarial question}
        \label{fig:adv_ex}
    \end{minipage}
\end{figure}

The review of the 199 episodic memories generated by GPT-4o-mini highlights different types of errors that the model identifies in its reflections. In 68 cases, the model correctly did not identify any error.
In \textit{Omission Errors}, 47\% of mistakes, the model failed to include relevant context or information, but still gave a partially correct answer;
    for errors under  \textit{Misinterpretation of Temporal Context}, 27\% of errors identified, the model did not correctly reason about dates; these correspond to almost all temporal questions in the sample.
   \textit{Assumption and Overgeneralisation} errors, 11\% of the total, include incorrect assumptions and broad conclusions drawn without evidence, usually about adversarial questions;
    in errors due to \textit{Lack of Specificity}, the model did not provide detailed and precise information, but typically still had a partially correct answer;
    errors due to \textit{Inclusion of Extraneous Information} included irrelevant details that might confuse the main point.
We report an example of reflection on temporal reasoning in Figure \ref{fig:ep_ex}.

\subsection{Impact of Procedural Memory}
The addition of procedural memory to RAG and episodic memory shows negative results in the LoCoMo QA setting. 
This outcome can be interpreted as reflecting a mismatch between the nature of the task and our current implementation of procedural memory. Procedural memory is intended to encode strategies and behaviours and is likely to become more effective for tasks requiring planning steps, multi-step reasoning, or tool usage rather than QA over conversations. 
On the other hand, on QMSum, RAG+PromptOpt achieves notable improvements for Llama (20.45 vs. 8.88 for baseline RAG), Qwen (23.32 vs. 20.94), and GPT (21.78 vs. 20.03), demonstrating that optimization techniques can benefit transfer learning:  our initial prompt asks for concise answers, while the QMSum target summaries tend to be more detailed and lengthy, so models optimize their prompts from experience to allow that.

Our qualitative analysis of the prompts generated for LoCoMo reveals that newly generated prompts often focus too closely on specific examples instead of generalizing behavioural patterns and tend to repeat instructions (Appendix \ref{sec:appendixb6}). This approach appears to suffer from overfitting to the training examples, creating prompts that are too narrow for the diverse reasoning required in our evaluation set.
This suggests that the optimization procedure captured surface-level corrections (e.g., succeeding in adversarial questions) rather than abstracting broader strategies applicable across contexts.

\subsection{Token Efficiency}
A key advantage of memory and retrieval augmentation approaches lies in their substantial reduction in token usage.
Full Context approaches consistently require between 23000 and 26000 tokens per query, RAG approaches with top-10 utterances and episodic or procedural memory use between 600 and 1500 tokens, A-Mem uses between 2300 and  3300 tokens per query -- not including the ones needed to populate the memory. 
This represents a reduction of token consumption of more than 90\% for RAG-based methods, with minimal or no degradation in overall performance. In particular, RAG+EpMem seems to offer the best balance of performance and efficiency for low-resource environments.

\section{Conclusions and future work}
This work provides an evaluation of memory augmentation strategies for long-context and knowledge-intensive tasks. We compare simple and complex implementations of semantic memory, and minimal implementations of episodic and procedural memory across multiple LLMs.  By studying semantic, episodic, and procedural memory under a unified framework, we identify distinct trade-offs between performance, efficiency, and interpretability.
Our findings confirm that while full-context prompting serves as a strong baseline, this approach suffers from inefficiency, poor scalability, limited interpretability, and vulnerability to context-length issues that can degrade performance.
Most foundation and instruction-tuned models can benefit from simple RAG approaches, but instruction-tuned models can make use of their improved instruction-following and reasoning capabilities and also show strong performance with more complex approaches to semantic memory like A-Mem and episodic memory integration. In particular, GPT-4o mini shows the best results with RAG+EpMem, highlighting the potential of experiential learning with textual feedback.

For knowledge-intensive and long-context tasks, semantic memory is essential, but the best implementation should be determined based on use cases, scalability requirements, efficiency constraints, and data structure. Our findings show that, for less powerful models, complex approaches like A-Mem do not work at the best of their potential. Additionally, though more efficient than full-context prompting at inference time, A-Mem is more resource-intensive than RAG to construct, it remains effective for stable data that can benefit from clustering, but would become inefficient when memory requires frequent updates.
Procedural memory, in this implementation, appears less suited to QA tasks but may prove valuable for planning or tool-use scenarios. Additionally, the results on QMSum show that it can help with transfer learning.
In alignment with previous findings  \cite{shinn2023reflexion, zhao2024expel}, episodic memory seems particularly valuable in enabling models to learn from both positive and negative experiences for complex QA tasks such as the adversarial one.  

Taken together, our results show the impact of different types of memory-augmentation for knowledge-intensive tasks in smaller-scale LLMs. At the same time, they highlight  directions for future research: extending evaluation to multi-turn coherence and developing selective, adaptive, metacognitively informed memory mechanisms. Future works should investigate agentic approaches to selecting experience examples, removing the need for fixed update schedules, and how the selection of positive or negative examples for episodic and procedural memory affects overall results. Comparing such non-parametric, memory-based adaptation with reinforcement learning methods may further illuminate how LLMs can learn not just from data, but from their own experiences.

\section*{Limitations}
Our study has several limitations that open up directions for future research.
Firstly, we don't study multi-turn QA directly, but it's worth noting that the dataset includes several question categories that test similar skills: multi-hop questions require connecting information across multiple conversation turns; temporal questions require understanding the chronological flow of conversations; adversarial questions test whether systems can distinguish between information that exists versus doesn't exist in the conversation history. 
Additionally, our experiments focus on small open-weight models ($\leq$7B parameters) and on an efficient commercial reasoning model (GPT-4o mini). This choice reflects realistic deployment constraints and enables reproducibility, and the results on GPT-4o mini validate our assumptions. Nonetheless, much larger models may exhibit different memory behaviours.
Finally, our implementations of episodic and procedural memory approaches are deliberately minimal, to isolate their effects, and we store memories only from the first sample out of 10 in the dataset. This artificial constraint limits both the benefits and complexities of a system where memories accumulate over time and throughout conversations. 
Our study did not address long-term memory management issues such as mechanisms for forgetting, updating, or consolidating memories and handling contradictions.
These limitations suggest directions for future research to develop more robust, efficient, and reliable memory-augmented conversational systems.

\begin{ack}
This research was partially funded by the UKRI AI Centre for Doctoral Training in Responsible and Trustworthy in-the-world Natural Language Processing (grant ref: EP/Y030656/1).
\end{ack}

\bibliographystyle{plain}
\bibliography{custom.bib}


\appendix

\section{Hyperparameter Settings}
For all answer-generation LLM calls and A-Mem we set the temperature to 0.5, to balance model creativity and accuracy. An higher temperature of 0.7 is used for reflection and optimized prompt generation steps, while lower (0.4) temperature is used for the prompt classification step, as the model should stick to generating a list of prompt names from the given one.

For the LoCoMo dataset, for RAG we decided to retrieve the top-10 most relevant utterances to both provide the model with enough relevant information and keep the context short, without introducing noise through the retrieval of less relevant utterances. For QMSum, we set $k=20$ due to the wider context summarisation queries required. Table \ref{tab:rag} shows the results with $k$ set to 5, 10, and 20 for Llama 3.2 3B. $K$ also aligns with the number of memory notes being retrieved by A-Mem.

\begin{table}[h!]
    \centering
        \caption{\textbf{Performance comparison of RAG with different $top-k$ on LoCoMo}. F1 scores, rankings and tokens per query across five categories: Single-Hop (S-Hop), Multi-Hop (M-Hop), Temporal (Temp.), Open-Domain (Open), and Adversarial (Adv.). Results from Llama 3.2 3B, best in bold.} 
    \scriptsize  
    \setlength{\tabcolsep}{3pt}  
    \renewcommand{\arraystretch}{0.85}  
    \begin{tabular}{@{}>{\raggedright\arraybackslash}p{0.05\linewidth}rrrrrrr@{}}
    \toprule
    \textbf{Top-k} & \multicolumn{5}{c}{\textbf{Category}} & \multicolumn{2}{c@{}}{\textbf{Average}} \\
    \cmidrule(lr){2-6} \cmidrule(l){7-8}
    & \textbf{S-Hop} & \textbf{M-Hop} & \textbf{Temp.}& \textbf{Open}& \textbf{Adv.}& \textbf{F1 Rank}& \textbf{Tokens} \\
    \midrule
    \midrule
         5 & 13.45	&\textbf{11.13}	&7.36	&13.03	&\textbf{4.48}& \textbf{1.6}& 384.90\\
         10 & 11.99	&10.64	&7.28	&\textbf{13.85}	&2.91& 2.2& 658.11\\
         20 & \textbf{14.74}	&8.64	&\textbf{7.53}	&12.78	&2.47& 2.2& 1196.08\\
    \bottomrule
    \end{tabular}
    \label{tab:rag}
\end{table}

\section{Prompts Used}
\label{sec:appendix}

\subsection{Full Context and LoCoMo Summaries}
\begin{Verbatim}[fontsize=\small, breaklines=true]
Based on the given conversations, write a short answer for the following question in a few words. Do not write complete and lengthy sentences. Answer with exact words from the conversations whenever possible.

Below is a conversation between two people: {name of speaker 1} and {name of speaker 2}. The conversation takes place over multiple days and the date of each conversation is written at the beginning of the conversation.

{full conversation transcript or session summaries}

If the answer to the question requires you to do temporal reasoning, use DATE of CONVERSATION to answer with an approximate date. If the question cannot be answered, write 'No information available.
Question: {question} Answer:
\end{Verbatim}

\subsection{RAG}
\begin{Verbatim}[fontsize=\small, breaklines=true]
Based on the given conversations, write a short answer for the following question in a few words. Do not write complete and lengthy sentences. Answer with exact words from the conversations whenever possible.

Below are retrieved snippets from a conversation between two people: {name of speaker 1} and {name of speaker 2}. 

{top-10 retrieved conversations snippets in the format 
"speaker: utterance, date of conversation: day month year"}

If the answer to the question requires you to do temporal reasoning, use DATE of CONVERSATION to answer with an approximate date. If the question cannot be answered, write 'No information available.
Question: {question} Answer:
\end{Verbatim}

\subsection{A-Mem}
\begin{Verbatim}[fontsize=\small, breaklines=true]
Based on the context: {top-10 retrieved memories containing talk start time, speaker and utterance, memory context, memory keywords, memory tags}, write an answer in the form of a short phrase for the following question. Answer with exact words from the context whenever possible. If the answer to the question requires you to do temporal reasoning, use DATE of CONVERSATION to answer with an approximate date. If the question cannot be answered, write 'No information available'.

Question: {question} Short answer:
\end{Verbatim}

\subsection{EpMem: Episodic Memory}
\subsubsection{Reflection Step}
\begin{Verbatim}[fontsize=\small, breaklines=true]
Reflect on your performance in the following QA example. Focus specifically on:
- Whether your answer was correct or not, and why.
- If the question required temporal reasoning, how well you handled it.
- If the question had no answer in the context, whether you correctly identified that.
- What reasoning errors (if any) occurred, and how to avoid them in future similar cases.
Provide a short reflection in a few sentences.

Question: {question}
Context: {top-10 retrieved utterances}
Correct Answer: {label}
Predicted Answer: {prediction}
Reflection:
\end{Verbatim}

\subsubsection{QA Step}
\begin{Verbatim}[fontsize=\small, breaklines=true]
You are an AI assistant that answers questions based on a given conversation. Use the current context and optionally refer to past examples and reflections to help you reason, but focus on the current question.
Based on the given conversation snippets, write a short answer for the following question in a few words. Do not write complete and lengthy sentences. Answer with exact words from the conversations whenever possible. Use the following examples to answer correctly. 

Examples: {top-3 most similar episodic memories with reflections}

Below are retrieved snippets from a conversation between two people.
{top-10 retrieved utterances}

Question: {query}

Now, answer the current question using the conversation context above. Refer to the past examples and reflections only if they help clarify your reasoning. If the answer to the question requires you to do temporal reasoning, use DATE of CONVERSATION to answer with an approximate date. If the question cannot be answered, write 'No information available'. Answer:
\end{Verbatim}

\subsection{Examples of Reflection}
\begin{Verbatim}[fontsize=\small, breaklines=true]
"question": "Who supports Caroline when she has a negative experience?",
"context": {top-10 retrieved utterances},
"answer": "Her mentors, family, and friends"
"prediction": " Melanie, friends, others. Note: The answer should be a few words as requested.",
"f1_score": 0.333,
"reflection": " I incorrectly identified Melanie as the primary support person for Caroline. While Melanie does offer support, the context of the conversations reveals that Caroline's support system is diverse and includes her mentors, family, and friends. I struggled with temporal reasoning in this case, as the question requires me to analyze the conversations across different dates to identify patterns and relationships. In future, I will be more careful to consider the broader context and not rely solely on individual conversations. Also, I should have noticed that the question was asking about a specific scenario (negative experience) rather than general support. This would have led me to focus on conversations that mention Caroline's negative experiences, such as her tough breakup and the challenges she faced."

"question": "Would Caroline likely have Dr. Seuss books on her bookshelf?"
"context": "Caroline: I've got lots of kids' books - classics, stories from different cultures, educational books, all of that. What's a favorite book you remember from your childhood?"...{other retrieved snippets}
"answer": "Yes, since she collects classic children's books",
"prediction": "No information available.",
"f1_score": 0,
"reflection": "In this case, I didn't correctly identify that Caroline's love of reading and collecting classic children's books would imply that she would likely have Dr. Seuss books on her bookshelf. I should have used the context of their conversation to infer this, rather than relying on my initial instinct. I will make sure to pay closer attention to the conversation flow and use the context to guide my answers in the future."
\end{Verbatim}

\subsection{PromptOpt: Procedural Memory}
\subsubsection{Original Prompts to Optimize}
\begin{Verbatim}[fontsize=\small, breaklines=true]
{
    "name": "task",
    "prompt": "Based on the above conversations, write a short answer for the following question in a few words. Do not write complete and lengthy sentences. Answer with exact words from the conversations whenever possible.",
},
{
    "name": "intro",
    "prompt": "Below are retrieved snippets from a conversation between two people. \n",
},
{
    "name": "rules",
    "prompt": "If the answer to the question requires you to do temporal reasoning, use DATE of CONVERSATION to answer with an approximate date. If the question cannot be answered, write 'No information available'.",
}
\end{Verbatim}

\subsubsection{Classification Step}
\begin{Verbatim}[fontsize=\small, breaklines=true]
You always return JSON output. Analyze the following trajectories and decide which prompts ought to be updated to improve the performance on future trajectories:

{batch of 5 LLM trajectories, containing query, predicted answer, correct answer, and F1}

Below are the prompts being optimized: {dictionary containing name of prompt: prompt content}

Return one JSON dictionary in the format {"which": [...]}, listing the names of prompts that need updates. The names must be in {prompt_names}. Do not return any explanations or reasoning. 
\end{Verbatim}

\subsubsection{Optimization Step}
\begin{Verbatim}[fontsize=\small, breaklines=true]
You are helping an AI assistant learn by optimizing its prompt. You always return JSON output.

## Background

Below is the current prompt: {prompt}

## Instructions

The developer provided these instructions regarding when/how to update:

<update_instructions>Do not make the prompts specific about any particular people or events mentioned in any question or conversation.<update_instructions>

## Session Data
Analyze the session(s) (and any user feedback) below:

<trajectories>{batch of 5 LLM trajectories, containing query, predicted answer, correct answer, and F1}<trajectories>

## Instructions

1. Reflect on the agent's performance on the given session(s) and identify any real failure modes (e.g., style mismatch, unclear or incomplete instructions, flawed reasoning, etc.).
2. Recommend the minimal changes necessary to address any real failures. If the prompt performs perfectly, simply respond rewriting the original prompt without making any changes.
3. DO NOT use any tags like <current_prompt>, <current_prompt> or <trajectories> in your response.
4. Be brief and concise. Avoid unnecessary verbosity.
IFF changes are warranted, focus on actionable edits. Be concrete. Edits should be appropriate for the identified failure modes. For example, consider clarifying the style or decision boundaries, or adding or modifying explicit instructions for conditionals, rules, or logic fixes; or provide step-by-step reasoning guidelines for multi-step logic problems if the model is failing to reason appropriately.
ONLY return JSON in the following format: {"reasoning": "<reasoning>", "updated_prompt": "<updated_prompt>"}.
\end{Verbatim}

\subsubsection{QA Step}
\begin{Verbatim}[fontsize=\small, breaklines=true]
{(optimized) task prompt}

{(optimized) intro prompt}
{top-10 retrieved utterances}

{(optimized) rules prompt}
Question: {query} Answer:
\end{Verbatim}

\subsubsection{Example of Optimized Prompts}
\label{sec:appendixb6}
The following prompt is the result obtained after optimization with Qwen2.5 7B Instruct:
\begin{Verbatim}[fontsize=\small, breaklines=true]
Based on the above conversations, write a short answer for the following question in a few words. Use exact words from the conversations whenever possible. If no exact words are available, provide a concise summary using only the information from the conversations, but avoid adding extra details or interpretations. Always confirm with 'No information available' if the required information is not present in the conversations. Only use 'No information available' when no relevant information is present.

Below are retrieved snippets from a conversation between two people.
{top-10 retrieved utterances}

Provide direct answers based on the given information only. If the answer can be extracted verbatim, do so. If no information is available, write 'No information available'. Avoid adding extra context, personal speculations, or any assumptions and ensure the answer is accurate. Always refer back to the provided snippets for answers. Only provide 'No information available' unless explicitly asked to infer or provide additional context.
Question: {query} Answer:
\end{Verbatim}

\newgeometry{bottom=1in}
\section{Full Results Table}
\label{sec:results}
\begin{table*}[th!]
   \caption{\textbf{Performance comparison of memory augmentation approaches across various language models and reasoning categories.} We report average F1 scores, rankings and average number of tokens used per query.} 
    \centering
    \scriptsize
    \setlength{\tabcolsep}{4pt}
    \renewcommand{\arraystretch}{0.9}
    \begin{tabular}{@{}>{\raggedright\arraybackslash}p{0.12\linewidth}lrrrrrrr@{}}
    \toprule
    \textbf{Model} & \textbf{Approach} & \multicolumn{5}{c}{\textbf{Category}} & \multicolumn{2}{c@{}}{\textbf{Average}} \\
    \cmidrule(lr){3-7} \cmidrule(l){8-9}
    & & \textbf{Single-Hop} & \textbf{Multi-Hop} & \textbf{Temp.}& \textbf{Open Dom.}& \textbf{Adv.}& \textbf{F1 Rank.}& \textbf{Tokens} \\
    \midrule
    \midrule
    \multirow{8}{*}{Llama 3B Inst.}& Full Context & \textbf{30.07}& \textbf{25.96}& \textbf{11.39}& \textbf{39.36}& 26.90& \textbf{1.00}& 23265.98\\
        & RAG& 5.96& 8.73& 4.29& 5.18& 17.26& 4.50& 658.11\\
        & A-Mem & 3.55& 4.31& 5.12& 4.04& 50.56& 4.00& 2480.37\\
        & RAG+PromptOpt& 10.51& 10.16& 6.88& 8.31& 33.08& 2.83& 821.42\\
        & RAG+EpMem& 19.96& 11.80& 5.59& 13.33& 67.71& 1.83& 1305.82\\
        & RAG+PromptOpt+EpMem& 17.40& 9.78& 3.85& 10.78& \textbf{72.18}& 2.83& 1377.53\\
    \midrule
    \multirow{8}{*}{Qwen 7B Inst.}& Full Context & 10.61& 3.39& 3.72& 8.24& 78.02& 4.50& 23376.87\\
        & RAG& \textbf{21.72}& 8.05& 4.89& \textbf{14.51}& 94.84& \textbf{2.00}& 695.74\\
        & A-Mem & 12.73& \textbf{8.67}& \textbf{6.28}& 13.86& 90.53& \textbf{2.00}& 3307.27\\
        & RAG+PromptOpt& 7.99& 1.97& 4.00& 9.11& 73.18& 4.66& 1024.88\\
        & RAG+EpMem& 17.51& 6.59& 5.35& 10.21& \textbf{95.24}& \textbf{2.00}& 1452.88\\
        & RAG+PromptOpt+EpMem& 17.64& 6.35& 5.33& 9.96& 94.48& 2.66& 1453.28\\
    \midrule
    \multirow{8}{*}{GPT-4o mini}& Full Context & 31.68& 20.40& 12.04& \textbf{56.40}& 52.13& 2.66& 23132.49\\
        & RAG& 29.98& 30.06& 10.13& 49.47& 84.46& 2.66& 649.17\\
        & A-Mem & 24.83& 26.02& 7.61& 37.48& 64.79& 3.83& 3514.45\\
        & RAG+PromptOpt& 15.58& 13.22& 5.96& 22.98& \textbf{92.98}& 4.00& 668.90\\
        & RAG+EpMem& \textbf{31.77}& \textbf{40.39}& \textbf{12.51}& 51.78& 77.69& \textbf{1.66}& 969.26\\
        & RAG+PromptOpt+EpMem& 30.79& 41.44& 5.86& 51.71& 80.53& 2.66& 972.93\\
    \midrule
    \midrule
     \multirow{8}{*}{Llama 3B}& Full Context & \textbf{14.72}& 9.78& 7.57& \textbf{14.74}& 0.67& \textbf{2.00}& 23265.98\\
        & RAG& 11.99& \textbf{10.64}& 7.28& 13.85& 2.91& 2.16& 658.11\\
        & A-Mem & 5.49& 4.11& \textbf{8.44}& 6.51& \textbf{11.36}& 3.33& 2309.74\\
        & RAG+PromptOpt& 11.47& 5.83& 6.53& 9.74& 6.26& 3.16& 657.42\\
        & RAG+EpMem& 9.15& 7.21& 4.57& 7.57& 10.27& 3.50& 1243.96\\
        & RAG+PromptOpt+EpMem& 9.72& 7.77& 4.44& 7.64& 8.74& 3.33& 1232.17\\
    \midrule
    \multirow{8}{*}{Qwen 7B}& Full Context & 16.96& 12.32& 8.17& 28.43& 5.17& 2.33& 23376.87\\
        & RAG& \textbf{21.50}& \textbf{23.14}& \textbf{9.48}& \textbf{31.32}& 44.84& \textbf{1.16}& 695.74\\
        & A-Mem &10.93& 9.30& 5.90& 14.33& \textbf{66.14}& 3.16& 2460.99\\
        & RAG+PromptOpt& 12.30& 8.29& 7.36& 12.98& 7.02& 3.50& 886.8\\
        & RAG+EpMem& 15.85& 7.32& 5.93& 12.72& 44.36& 3.83& 1304.66\\
        & RAG+PromptOpt+EpMem& 15.46& 7.42& 6.03& 12.53& 47.61& 3.50& 1402.7\\
    \bottomrule
    \end{tabular}
    \label{tab:memory_results_all}
\end{table*}


\end{document}